\documentclass[sigconf,nonacm]{acmart}

\usepackage{balance}
\usepackage{caption}
\usepackage{subcaption}
\usepackage{multirow}
\usepackage{graphicx}
\usepackage{gensymb}
\usepackage{soul}
\usepackage{amsmath}

\usepackage{xargs}                      
\usepackage[colorinlistoftodos,prependcaption,textsize=tiny]{todonotes}
\newcommandx{\unsure}[2][1=]{\todo[linecolor=red,backgroundcolor=red!25,bordercolor=red,#1]{#2}}
\newcommandx{\change}[2][1=]{\todo[linecolor=blue,backgroundcolor=blue!25,bordercolor=blue,#1]{#2}}
\newcommandx{\info}[2][1=]{\todo[linecolor=green,backgroundcolor=green!25,bordercolor=green,#1]{#2}}
\newcommandx{\improvement}[2][1=]{\todo[linecolor=cyan,backgroundcolor=cyan!25,bordercolor=cyan,#1]{#2}}
\newcommandx{\thiswillnotshow}[2][1=]{\todo[disable,#1]{#2}}

\DeclareMathOperator{\EX}{\mathbb{E}}

\usepackage{afterpage}
\usepackage{tabularray}
\DefTblrTemplate{firsthead,middlehead,lasthead}{default}{
}
\DefTblrTemplate{firstfoot}{default}{
}
\DefTblrTemplate{middlefoot}{default}{
}
\DefTblrTemplate{lastfoot}{default}{
  \UseTblrTemplate{note}{default}
  \UseTblrTemplate{remark}{default}
  \UseTblrTemplate{caption}{default}
}

\usepackage[normalem]{ulem}

\usepackage[version-1-compatibility]{siunitx}

\AtBeginDocument{%
  \providecommand\BibTeX{{%
    \normalfont B\kern-0.5em{\scshape i\kern-0.25em b}\kern-0.8em\TeX}}}

\setcopyright{acmcopyright}
\copyrightyear{2018}
\acmYear{2018}
\acmDOI{10.1145/1122445.1122456}

\acmConference[Woodstock '18]{Woodstock '18: ACM Symposium on Neural
  Gaze Detection}{June 03--05, 2018}{Woodstock, NY}
\acmBooktitle{Woodstock ’18: ACM Symposium on Neural Gaze Detection,
  June 03--05, 2018, Woodstock, NY}
\acmPrice{15.00}
\acmISBN{978-1-4503-XXXX-X/18/06}

\balance

\begin{document}


\title[Teach Me How to Learn]{Teach Me How to Learn: A Perspective Review towards User-centered Neuro-symbolic Learning for Robotic Surgical Systems}


\author{Amr Gomaa}
\affiliation{%
  \institution{German Research Center for Artificial Intelligence (DFKI)}
  \country{}
}
\affiliation{%
  \institution{Saarland Informatics Campus}
  \city{Saarbr{\"u}cken}
  \country{Germany}
}

\email{amr.gomaa@dfki.de}

\author{Bilal Mahdy}
\affiliation{%
  \institution{German Research Center for Artificial Intelligence (DFKI)}
    \city{Saarbr{\"u}cken}
  \country{Germany}
}

\email{bilal.mahdy@dfki.de}

\author{Niko Kleer}
\affiliation{%
  \institution{German Research Center for Artificial Intelligence (DFKI)}
    \city{Saarbr{\"u}cken}
  \country{Germany}
}
\email{niko.kleer@dfki.de}

\author{Michael Feld}
\affiliation{%
  \institution{German Research Center for Artificial Intelligence (DFKI)}
    \city{Saarbr{\"u}cken}
  \country{Germany}
}
\email{michael.feld@dfki.de}

\author{Frank Kirchner}
\affiliation{%
  \institution{German Research Center for Artificial Intelligence (DFKI)}
  \country{}
}
\affiliation{%
  \institution{Robotics Research Group at University of Bremen}
  \city{Bremen}
  \country{Germany}
}

\email{frank.kirchner@dfki.de}

\author{Antonio Kr{\"u}ger}
\affiliation{%
  \institution{German Research Center for Artificial Intelligence (DFKI)}
    \city{Saarbr{\"u}cken}
  \country{Germany}
}

\email{antonio.krueger@dfki.de}

\renewcommand{\shortauthors}{Gomaa et al.}

\begin{abstract}

Recent advances in machine learning models allowed robots to identify objects on a perceptual nonsymbolic level (e.g., through sensor fusion and natural language understanding). However, these primarily black-box learning models still lack interpretation and transferability and require high data and computational demand. An alternative solution is to teach a robot on both perceptual nonsymbolic and conceptual symbolic levels through hybrid neurosymbolic learning approaches with expert feedback (i.e., human-in-the-loop learning). This work proposes a concept for this user-centered hybrid learning paradigm that focuses on robotic surgical situations. While most recent research focused on hybrid learning for non-robotic and some generic robotic domains, little work focuses on surgical robotics. We survey this related research while focusing on human-in-the-loop surgical robotic systems. This evaluation highlights the most prominent solutions for autonomous surgical robots and the challenges surgeons face when interacting with these systems. Finally, we envision possible ways to address these challenges using online apprenticeship learning based on implicit and explicit feedback from expert surgeons.

\end{abstract}



\begin{teaserfigure}
  \centering\includegraphics[width=0.77\linewidth]{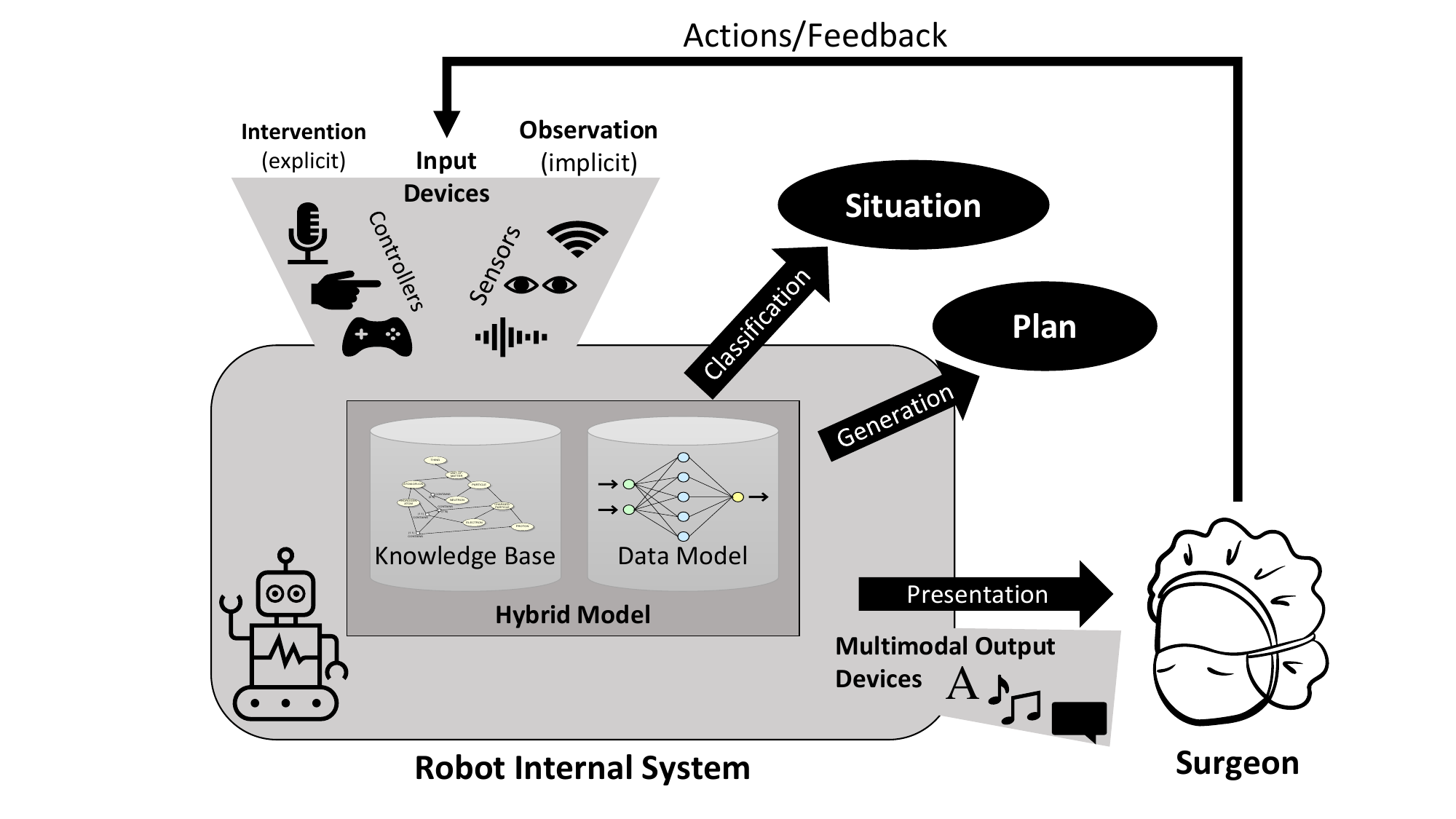}
  \caption{Overview of the envisioned user-centered neuro-symbolic human-in-the-loop learning system. The expert surgeon and the robotic agent are in a continuous feedback loop, starting with the surgeon's demonstrations (i.e., actions for different states), then judging the robot's output and providing support through thorough feedback.}
  \Description{This images shows an overview of the user-centered neuro-symbolic learning system where the surgeon expert can view the robot behaviour through mutlimodal output and give feedback with corrected actions through multimodal inputs.}
  \label{fig:teaser}
\end{teaserfigure}

\maketitle

\section{Introduction}

\textit{Apprenticeship} is defined as ``a system for training a new generation of practitioners of a trade or profession with on-the-job training''\footnote{\url{https://en.wikipedia.org/wiki/Apprenticeship}}; so imagine a system where a professional human worker can teach an industrial apprentice robot. Recent advances in machine learning and deep learning algorithms allowed robots to understand objects and the surrounding environment on a perceptual nonsymbolic level (e.g., object detection, sensor fusion, and language understanding); however, recent approaches attempted to understand objects on a conceptual, symbolic level to achieve more human-like learning for the robotic domain ~\cite{zhang2017learning,evans2018learning,liao2017object}. 

While subsymbolic learning methods, such as neural networks, have shown remarkable results in fields such as computer vision, NLP, and NLU, they still suffer from a lack of explainability, computationally exhausting, large dataset requirements, and limited transferability. On the other hand, while symbolic learning is ``legible'' by humans, it can lead to combinatorial growth, making solutions to complex problems unfeasible. When both types of learning are combined, it could be possible to obtain advantages while overcoming disadvantages. For example, a teacher might teach a robot how to tidy up a table full of bottles at different stages. In the first stage, the teacher might guide the robot arm, showing it how to remove one bottle from the table (subsymbolic learning by example). In the next stage, when the basic movements have been acquired, supervised learning can continue through verbal instructions (symbolic learning by instruction)~\cite{grumbach_learning_1995}. Similarly, the surgeon can guide a surgical robot first to learn basic surgical actions (like a surgical apprentice), then it can perform complete surgeries independently.

Several approaches have been proposed to insert human knowledge into neural networks to initialize, refine, guide, and speed up the learning process, then extract symbolic information from the network~\cite{shavlik_combining_1994,von2021informed}. However, most approaches are inapplicable to long-term sustainable autonomous systems such as automated surgical robots~\cite{garcez2020neurosymbolic,lecun2022path}.
Another implication of long-term autonomous systems is the need for efficient learning methods that suit the available training inputs. Promising techniques include active learning~\cite{Settles2009active}, inverse reinforcement learning~\cite{Hadfield-Menell2016nips}, interactive machine learning~\cite{Amershi2014}, and machine teaching~\cite{Simard2017}.
More recent attempts explicitly combined deep learning with knowledge bases in joint models (e.g., for construction and population)~\cite{ratnerAlexEtAl2018, adel2018deep} to ultimately achieve human-like performance. Some work has focused on integrating neural networks with classical planning by mapping sub-symbolic input to a symbolic one, which automatic planners can utilize~\cite{asai2018classical}. Others have used Logic Tensor Networks to enable learning from noisy data in the presence of logical constraints by combining low-level features with high-level concepts~\cite{serafini2016logic, donadello_logic_2017}. Other approaches included psychologically inspired cognitive architectures by having a goal-directed organizational hierarchy with parallel subsymbolic algorithms running in the lower levels and symbolic ones running serially in the higher levels \cite{kelley_developing_2006}. Furthermore, some approaches looked at the domain of generative adversarial networks (GANs~\cite{goodfellow2014generative}) domain, such as generative adversarial imitation learning (GAIL~\cite{ho2016generative}) and interpretable imitation learning (infoGAIL~\cite{li2017infogail}).
However, these approaches had several drawbacks, such as a highly prolonged training time and increased computational resources. Additionally, the environmental interaction was mainly in a simulated setting that lacks external validity (i.e., applicability to realistic settings).

Similarly, researchers recently focused on human-centered artificial intelligence (HCAI)~\cite{xu2019toward,nowak2018assessing,bryson2019society,shneiderman2020human} to improve system performance, as well as to promote trust, control, and reliability between users and machines~\cite{shneiderman2020human}. Some researchers focused on situation-adapting learning techniques such as relational graph neural networks~\cite{jing2020relational}, while others focused on learning from user implicit feedback, such as efficient learning from demonstrations~\cite{li2022efficient}, and the implicit human feedback learner~\cite{cui2020empathic}. However, implementing the concepts of HCAI is still an ambiguous and challenging task that is lacking in recent work and robotic manipulation tasks~\cite{nowak2018assessing,bryson2019society}. 

Thus, this work presents an application-oriented survey of these approaches with a focus on the human-in-the-loop learning aspect (i.e., HCAI). We highlight the feasibility of these approaches in the surgical robotic domain and propose design concepts to achieve a machine teaching system based on the apprenticeship learning paradigm. More specifically, the artificial agent (i.e., surgical robot) would explicitly learn at both perceptual and conceptual levels through direct feedback from a human teacher (i.e., expert surgeon) while learning and understanding implicitly using its current view (i.e., sensors) of the world in a natural multimodal manner (see~\autoref{fig:teaser}). We focus on human-in-the-loop techniques and algorithms where a human is available as a learning source, which can be leveraged using interactive machine learning, enabling efficient and continuous updates to the machine learning models. This is particularly useful for enhancing user interaction because it enables robust processing of versatile signal streams and joint data analysis from multiple modalities and sensors to understand complex users' behavior and actions.

\section{Background and Related Work}

This section summarizes relevant review papers that surveyed hybrid machine learning and artificial intelligence robotic systems. Afterward, we introduce the concept of automating surgeries using robotic systems. Finally, we present the research questions that guided this survey.

\subsection{Machine learning and AI robotic systems}

Several manuscripts have been published on neurosymbolic learning, machine learning, and AI robotic manipulation. Garcez et al.~\cite{garcez2022neural} surveyed articles integrating machine learning and automated reasoning to achieve a human-like level of understanding, while Harmelen and Teije~\cite{van2019boxology} defined a boxology for existing machine learning algorithms, including hybrid learning, neuro-symbolic learning, explainable AI, and automated reasoning. Similarly, Rueden et al.~\cite{von2021informed} created a taxonomy of algorithms that integrate prior knowledge into machine learning systems such as physics-guided neural networks~\cite{daw2017physics,stewart2017label,battaglia2016interaction} and graph neural networks~\cite{marino2017more,battaglia2018relational}. While these works highlighted several approaches for integrating prior knowledge into learning algorithms, they are mostly theoretical use-case-specific approaches that focus on perceptual-level learning and do not apply to real-time practical robotic systems. 

Furthermore, Li et al.~\cite{li2019human}, Hussein et al.\cite{hussein2017imitation}, and Arzate et al.~\cite{arzate2020survey} surveyed human-centered reinforcement learning, imitation learning, and interactive reinforcement learning, respectively, which is more applicable to a robotic environment; however, they did not focus on robotic scenarios. On the other hand, Hua et al.~\cite{hua2021learning}, Fang et al.~\cite{fang2019survey}, Ravichandar et al.~\cite{ravichandar2020recent}, and Lee~\cite{lee2017survey} focused on reinforcement learning (RL), imitation learning (IL), and learning from demonstration (LfD) approaches specifically for robotic manipulation. However, unlike our survey, these approaches neither focus on neurosymbolic learning approaches nor verifiable algorithms explicitly designed for robotic surgical systems that are highly critical in terms of safety and require rigorous algorithmic performance. Therefore, in this work, we survey work focusing on robotic surgical system automation with the guidance of our research questions explicitly designed for this task. 


Finally, Hoeckelmann et al.\cite{hoeckelmann2015current}, Attanasio et al.\cite{attanasio2021autonomy}, and Haidegger et al.\cite{haidegger2022robot} provide a summary of existing robotic surgical systems, including the da Vinci robot\cite{dimaio2011vinci, intuitive2022Davinci}, Curexo's ROBODOC~\cite{curexo2014robodoc} (now called TSolution One~\cite{ThinkSurgical2014Tsolutionone}), and other telesurgical robots used for robot-assisted minimally invasive surgery (RAMIS). However, these surveys emphasize that current systems offer only lower levels of autonomy and emphasize the need for further research to achieve higher levels of seamless interaction and autonomy. Therefore, in this perspective review, we focus on user-centered machine learning approaches that utilize human-in-the-loop techniques to promote trust and improve performance.

\subsection{Automation of Surgical Robots}

Surgery, as a medical specialty, uses operative manual and instrumental techniques applied to the human body to treat a patient’s injury or pathological condition. Such microscopic procedures are performed manually by experienced surgeons with the help of a surgical microscope and require, in addition to trained hand-eye coordination and the planned procedure, machines, and technology to make time-critical and risky interventions feasible.
Surgical robotic imaging microscopes are currently manually configured by the operating surgeon to ensure optimal support through the imaging properties of the visual device directed at the surgical zone. However, imaging requirements in terms of positioning, viewing, and focus plane may change throughout the operation, making it necessary for surgeons to adapt the configuration of the surgical microscope. Since adjustments have to be performed manually, a short-term interruption of the operation becomes necessary. Individual risk considerations are made in each case as to whether a visual section correction is effective in the present surgical situation, which could negatively affect the outcome of the surgery if taken incorrectly.
Hence, there have been recent advances in the hand-free configuration of the microscope, such as mouth-controlled joysticks or foot control with the help of pedals~\cite{holly1976mouth,sindou2009practical,afkari2014potentials}. However, these approaches are rather impractical and allow too few degrees of freedom to meet the complex requirements for the various settings of the microscopic frame.

\subsection{Guiding Research Questions}
We formulated some guiding research questions to define our scope to cover the essential obstacles and milestones achieved in AI-based robotic surgery. Our answers will address the relative performance differences between using machine learning approaches that do not utilize a human-in-the-loop approach and using those that do. Additionally, we investigate the social barriers to employing AI-based solutions in different areas of medicine, namely, diagnoses and surgical robotics. Finally, we discuss the various difficulties in designing a human-in-the-loop learning approach in a simulated surgical environment (SSE). Our five compiled questions are as follows.

\begin{itemize}
    \item[\textbf{RQ1:}] How do current machine learning approaches perform in SSE?
    \item[\textbf{RQ2:}] How is AI-based robotic surgery perceived by surgeons and patients?
    \item[\textbf{RQ3:}] How can current human-in-the-loop learning approaches be utilized to enhance performance and promote trust?
    \item[\textbf{RQ4:}] What kind of expert knowledge is relevant in human-in-the-loop scenarios for SSE?
    \item[\textbf{RQ5:}] Which components of a real surgery offer a significant improvement to robot training when transferred to SSE?
\end{itemize}

Although our research questions focus on aspects related to surgical robots, a thorough analysis of related machine learning and human-in-the-loop learning is needed. Therefore, we focus the first part of our surveying methodology on hybrid learning approaches as a background introduction before addressing the surgical robot aspect, as discussed further. 

\section{Methodology}

First, this section highlights approaches and algorithms relating to the hybrid (e.g., neuro-symbolic) learning concept. Next, we focus on general AI in surgery and robotic surgeons' automation. Finally, we present work related to machine learning from a human surgeon.

\subsection{Hybrid Learning Methods}

There are several approaches to learning from subsymbolic (i.e., perceptual) and symbolic (i.e., conceptual) data in general. We define three separate categories for this hybrid neurosymbolic learning approach as follows: \textit{Concept Learning} (i.e., Symbolic Networks), \textit{Logic Neural Networks}, and \textit{Hybrid Deep Reinforcement Learning}. While these approaches are discussed in detail in this section, we acknowledge that this list is not exhaustive, as the goal is to showcase relevant hybrid learning approaches to our specific use case rather than discuss all existing approaches for all domains.

\subsubsection{Concept Learning:}

Several researchers have investigated machine learning on an abstract conceptual level. Bottou~\cite{bottou2014machine} discussed a bottom-up approach to reach a level of machine reasoning by concatenating several simple modules of machine learning tasks to achieve a complicated goal. Zhang and Sornette~\cite{zhang2017learning} introduced the Deep Symbolic Networks (DSN) that unravel the black-box property of a traditional deep neural network by linking network predictions with symbolic meanings. For example, instead of just outputting a cat or dog in a binary classification approach, the algorithm would output graph-like symbols indicating a cat has whiskers or a dog has a more extended sticking-out tongue. Liao and Poggio~\cite{liao2017object} introduced the Object-Oriented Deep Learning approach, where learning occurs at the object level instead of feature-oriented approaches. They also primarily focused on the interpretation and explainability aspect through learning and outputting visual object properties such as position, pose, and scale. Similarly, Yi et al.~\cite{yi2018neural}, Mao et al.~\cite{Mao2019NeuroSymbolic}, and Han et al.~\cite{Han2019Visual} introduced the Neuro-Symbolic Concept Learner (NS-CL) and Visual Concept-Metaconcept Learning (VCML) that learn visual concepts on an abstract level and understand object properties such as color, material, and spatial location. They build their architecture through several learning modules utilizing Convolutional Neural Networks (CNNs), Reinforcement Learning (RLs), and Curriculum Learning. However, these previous approaches are mainly focused on interpretation and explainability and do not necessarily achieve higher performance than their subsymbolic counterparts. Moreover, they are only applicable to the visual recognition domain and are not directly transferable to the robotic domain without further modification of their approaches, as discussed later.

 \subsubsection{Logic Neural Networks:}

The use of neural networks in inductive learning and logic programming is a technique conceptualized more than 25 years ago. Garcez and Zaverucha~\cite{garcez1999connectionist} were the first to integrate prior knowledge into a neural network algorithm to create the Connectionist Inductive Learning and Logic Programming System (C-IL\textsuperscript{2}P). More recently, Serafini et al.~\cite{serafini2016logic} created the Logic Tensor Network (LTN) method, formulating problems into first-order logic and encapsulating the neural network within differentiable learnable axioms. They utilize their method for different applications such as image interpretation~\cite{donadello_logic_2017} and, most importantly, combining prior knowledge into reinforcement learning algorithms using the transfer learning technique~\cite{badreddine2019injecting}. While these approaches show great promise in the computer vision domain and might also be beneficial for the robotics domain, they rely on complicated data representation in abstract logic or first-order logic that is not directly applicable to complicated robotic tasks, specifically surgical tasks.

\subsubsection{Hybrid Deep Reinforcement Learning:}

\paragraph{Reinforcement Learning (RL)}

RL dates back to 1983 when it was first introduced in the research areas of cybernetics, statistics, psychology, and neuroscience~\cite{kaelbling1996reinforcement,kiumarsi2017optimal}. In 1996, Kaeblings et al.~\cite{kaelbling1996reinforcement} highlighted the use of RL in artificial intelligence (AI) research as a way of teaching an agent how to complete a task based only on reward and punishment (i.e., without specifying how the task is actually achieved). With recent advances in computational power, researchers were able to combine RL with deep learning approaches and apply these algorithms to specific tasks with a predefined reward system (e.g., Atari games~\cite{mnih2013playing}). However, due to the computationally expensive nature of RL, researchers could not apply such algorithms to complex tasks. Additionally, current advances in the RL approach depend on hand-crafted sensitive unstable reward systems that require high computational power and extended training time. Therefore, the need for a reward-free RL algorithm increased. 

\paragraph{Constrained Reinforcement Learning}

To overcome some of the reinforcement learning challenges, Garnelo et al.~\cite{garnelo2016towards} proposed adding an interpretable symbolic front end to the traditional reinforcement learning architecture to be able to verify the agent actions and decisions. They demonstrated their approach in a simple proof-of-concept prototype to achieve the desired goal. However, their approach lacked external validity and transferability to other domains. Additionally, their algorithm only solves the verification problem of RL, but does not solve the computationally exhaustive challenge. Consequently, Verma et al.~\cite{verma2018programmatically} proposed explicitly constraining the learning space of traditional RL by combining it with programmatic rules. They added a domain-specific policy to the update rule of the RL algorithm to limit the learning space and project the agent output (at each step) into a space of explainable actions. Moreover, they applied their algorithm to an autonomous driving simulation~\cite{wymann2000torcs} instead of simple game-like use cases. However, while their approach was designed to overcome most of the reinforcement learning challenges such as verification, interpretation, debugging, and commonsense knowledge (i.e., conceptual knowledge); their algorithm is still computationally exhaustive, domain restricted (i.e., low transferability), and requires complicated data engineering of expert knowledge.

\paragraph{Inverse Reinforcement Learning (IRL)}

IRL dates back to 2000 as an approach to overcome the reward shaping~\cite{laud2004reward} and complicated engineering problems of the previously mentioned RL approaches~\cite{ng2000algorithms}. Abbeel et al.~\cite{abbeel2004apprenticeship} introduced apprenticeship learning and the possibility of explicitly teaching artificial agents without the need for complicated reward formulation. Torabi et al.~\cite{torabi2018behavioral} suggested a behavioral cloning-based approach that observes all user actions for different environmental states and learns a pattern to follow for each state and action pair. However, these algorithms were still computationally expensive and nonpractical, as they depended on heuristic approaches and linear programming concepts to achieve their goal. They also relied on complete knowledge of the environment (i.e., fully observable environment), which does not apply to realistic, partially observable settings. 

\paragraph{Generative Adversarial Imitation Learning}

Recently, Ho et al.~\cite{ho2016generative} suggested generative adversarial imitation learning (GAIL) that would overcome previous problems by learning from an expert demonstration instead of a trial and error that is applied in RL and IRL. This is achieved through a generative adversarial network (GAN)~\cite{goodfellow2014generative} architecture where the agent ``Generator'' must generate actions as close to the human expert as possible to be indistinguishable from the human action when judged by another part of the network known as the ``Discriminator''. Subsequently, Li et al.~\cite{li2017infogail} extended GAIL into infoGAIL as an optimized solution that can be used explicitly for visual data and produce more human-like actions. While these approaches show great promise in standard RL benchmarks (e.g., Atari games and driving simulations) and simple robotic tasks (e.g., pick-and-place task~\cite{lozano1989task}), they are not entirely applied to complicated real-time online procedures as envisioned in this perspective review.


\subsection{Artificial Intelligence and Surgical Robots}

Surgical robot automation using AI has been generally attempted in previous work~\cite{baek2018path,richter2019open,barnoy2021robotic}. However, these approaches focus either on the path planning of the surgical robotic arm or the sensing part of the cutting tool attached to it. Thus, these approaches are not directly applicable to fully automated surgical robots that rely on computer vision. Recently, Wagner et al.~\cite{wagner2021learning} attempted an image-based approach to teach a robot how to perform minimally invasive surgery. However, this approach was trained on a small dataset, and it was also trained with a basic machine learning algorithm (i.e., decision trees) that is lacking in performance and inapplicable to realistic settings. The authors also mention that while the use of an RL approach generally shows promise in robotics, it still poses a severe safety issue in the medical domain. 

\subsubsection{Reinforcement learning in a simulated surgical environment (SSE):}

In a review conducted by Datta et al.~\cite{DATTA2021329}, the use of reinforcement learning in healthcare (more specifically, surgery) was surveyed in terms of performance, achievements and challenges that hinder its progress. Datta et al. presented a summary of relevant achievements and significant progressive findings in surgical decision-making scenarios based on reinforcement learning. However, they also highlight the two major setbacks that must be faced in the reinforcement learning model: \textit{Patient State Determination} and \textit{Reward Formulation}. 

\paragraph{Agent State Determination}
It defines what information should be fed to the agent to trigger an action. In real-life scenarios, surgeons are exposed to an abundance of information that they filter through their knowledge and experience and act accordingly. However, exposing an RL agent to this much information is unrealistic and computationally expensive. This presents the first challenge in introducing RL to SSE: Choosing the right amount of information for the state/action pairs to provide the agent without overwhelming it with unnecessary information or underwhelming its performance with too little information.

\paragraph{System Reward Formulation}
After being provided with the environment state, the agent has to rate its own performance by scoring the change from the previous state to the current state based on its previous action. This occurs through a reward function that incentivizes the agent to perform better over time. The second challenge lies in formulating a reward function for a highly complicated scenario, such as robotic surgery, which can be extremely difficult.

\subsubsection{Current Approaches:}
While existing tools bridge the gap between reinforcement learning algorithms and robotic surgery simulations such as dVRL~\cite{dVRL}, methodologies that are concluded with positive results usually restrict the use case to a specific narrowed part of the surgical environment. Datta et al.~\cite{DATTA2021329} restrict the scope of the reviewed reinforcement learning scenarios to helper tasks during surgery protocols, such as determining an IV dose or monitoring patient-assisted ventilation machinery. On the other hand, more 'hands-on' use cases usually fall short when the environment becomes too complex for the agent to handle. Bourdillon et al.~\cite{rl_virtual_robotic_surgery} combined RL in a virtual robotic surgery simulation environment to train a proximal policy optimization (PPO) agent to operate a pair of scissors on a tissue-like object. The reliability of the results was positive when motion was restricted to a single axis, but dropped significantly when more than one was introduced. Other applications of reinforcement learning in surgical procedures such as~\cite{nguyen_pinch,su_multicamera,barnoy_lean_rl,ning_ultrasound,baek_pathplanning,gao_gesture_recognition,xu_surrol} are also similarly bound by complexity constraints or limited function.

\begin{figure*}[t]
    \centering
    \includegraphics[width=0.8\linewidth]{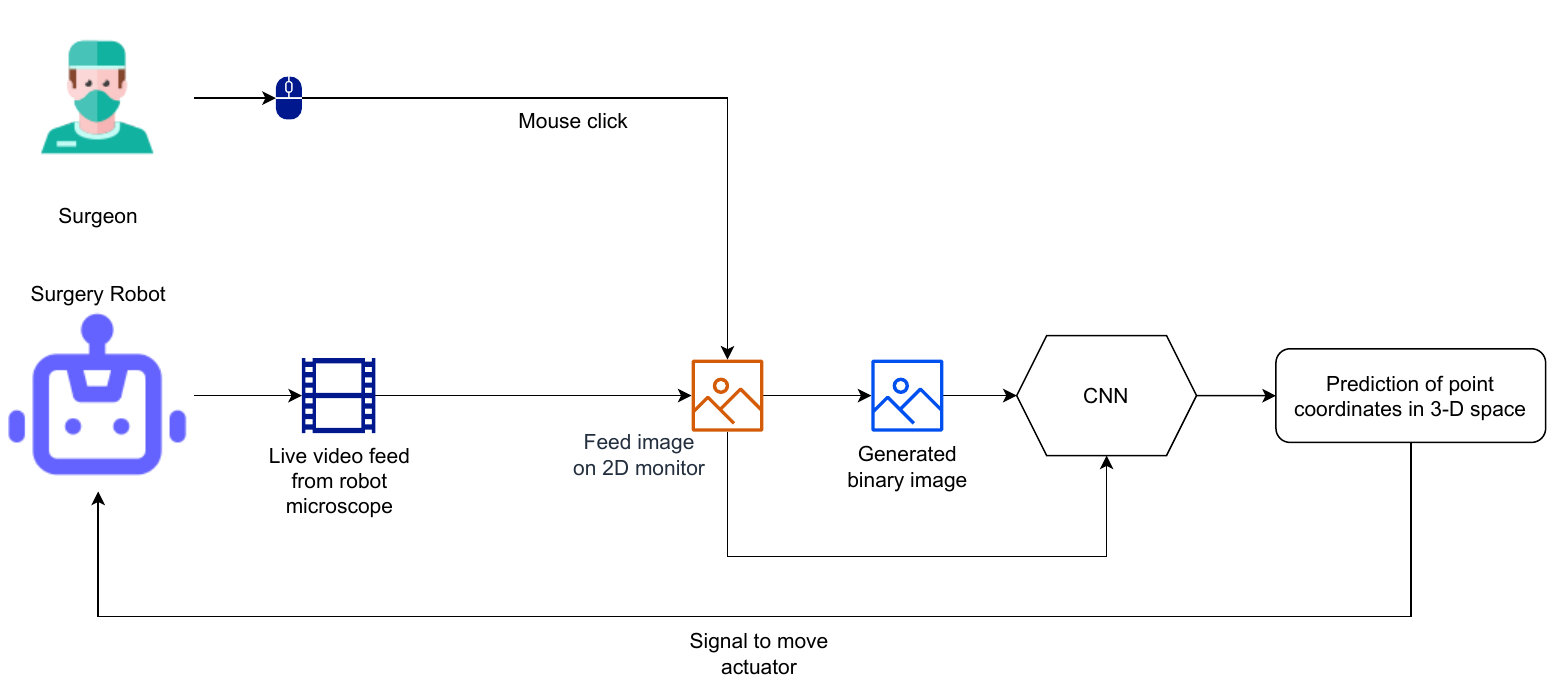}
    \caption{A high-level view of the pipeline implemented by Kim et al.~\cite{Kim2021_il}. The surgeon clicks on a particular point on the feed image, generating a binary image that represents the selected point. A CNN is then used to translate these images into a point in the 3D space to which the surgical robot actuators have to move.}
    \label{fig:johns_hopkins_flow_chart}
    \Description{Johns Hopkins High-level Pipeline where the surgeon clicks on a particular point on the feed image, which then generates a binary image depicting the selected point. A CNN is then used to translate these images into a point in 3D space to which the surgery robot actuators have to move. }
\end{figure*}

\subsubsection{Surgeon and Patient Perception:}

In 2015, a survey was published in the Surgical Endoscopy Journal~\cite{boys2016public} on the public perception of robot-assisted surgery, the surgeons who use them and the hospitals that deploy them. It showed that 67\% of the participants (n = 747) were concerned that a robot-assisted surgery would cause significant failure and cause internal damage to the patient. However, the same survey showed that around of 50\% of them did not understand correctly what a robot-assisted surgery is and half of those participants imagined that a robot-assisted surgery would be entirely autonomous. While this survey highlights the need for increasing the awareness for robot-assisted systems among patients, it further shows that patient trust would increase when having a real-time feedback-based learning algorithm where the surgeon can teach the robotic apprentice and stop it from making mistakes in an online manner.
On the other hand, a more recent survey was published in the Journal of Endourology~\cite{Stai2020-va} in 2020, in which participants were randomly assigned one of two questionnaires: one depicting a diagnosis made by a human physician and the second by an artificial intelligence (AI) application. The included topics covered several application areas for AI in medicine, such as diagnostics, prognostics, and surgical robotics. In contrast to the previous survey, participants are unaware of the origin of the diagnosis and therefore would be less influenced by any misconception about AI or robot-assisted surgery.
The results show that the participants (n = 264) had nearly the same trust in AI vs. physician diagnoses. However, they were significantly more likely to trust an AI diagnosis of cancer over a doctor diagnosis when responding to the version of the survey that suggested that an AI could make medical diagnoses (p = 9.32e-06). Although 55\% of the respondents (n = 145) reported that they were uncomfortable with automated robotic surgery, most of the individuals surveyed (88\%) mistakenly believed that partially autonomous surgery was already occurring. Almost all participants (94\%, n = 249) stated that they would be willing to pay for a review of medical imaging by an AI if available. This emphasizes the idea that healthcare providers should be aware of the potential for misinformation and the sensitivity of patients to how AI technology is represented as its applications increase in medical practice. Regarding surgeon perception of robotic surgery, Randell et al.~\cite{randell2015impact} investigated the point of view of the surgical team and reported the benefits of these systems such as reduced distraction, alleviation of surgeon stress and fatigue, and general relaxation. However, they also reported several challenges, such as reduced situational awareness since the surgeon is focused on a small part of the surgery, as well as lack of immersion and tactile feedback. Thus, the authors suggest further training and positive communication strategies among the entire team, including the robotic assistant. As suggested in this survey, this can be achieved through a user-centered surgeon-in-the-loop learning approach that will directly promote trust and positive communication due to the immediate feedback capabilities.

\subsection{Surgeon-in-the-loop Learning}

The previously mentioned restriction of complexity present in current robotic surgery applications becomes less prominent when imitation learning is utilized instead of traditional reinforcement learning. 

\subsubsection{Introducing expert demonstrations:}
In an experiment conducted by Johns Hopkins University~\cite{Kim2021_il} in which tool manipulation is automated by an agent trained using imitation learning, the simulation aimed to replicate an eye surgery environment in which precise tool movement around the delicate retina tissue is a significant factor in the performance outcome of the surgery. The proposed framework performed accurate navigation with an X-axis error of 0.089mm and a Y-axis error of 0.118mm, which is less than the human surgeon's mean tremor at the tooltip of 0.180mm for both axes. \autoref{fig:johns_hopkins_flow_chart} shows a high-level flow diagram of the methodology used by Kim et al.~\cite{Kim2021_il}.

\subsubsection{Adversarial solutions:}
Highly-capable imitation learning algorithms such as GAIL~\cite{ho2016generative} also help alleviate the complexity barrier in robotic surgery. This is shown in multiple proposals such as~\cite{tan_laparoscopy, li_laparoscopy,chi_catheter, segato_keyhole}. Tan et al.~\cite{tan_laparoscopy} proposed a training system for the Laparoscopy surgical procedure designed to improve surgeons' practical skills through online feedback, as well as demonstrations of prior knowledge and agent training. The system uses proximal policy optimization (PPO) to learn the control policy in the simulation and a generative adversarial imitation learning agent trained on both the learned policy and the prior knowledge of expert surgeons. Then it generates demonstration policies on robot-assisted devices for trainees and produces feedback from the training results.

In a different surgical domain, Segato et al.~\cite{segato_keyhole} employ a combination of PPO and GAIL to plan an insertion path for a steerable needle in Keyhole Neurosurgery. Being a minimally invasive surgery, the Remote Center of Mass (RCM) constraint is introduced, increasing the complexity of the problem. The agent is trained and tested in a simulated environment developed with \textit{Unity Game Engine} and then adapted to the Robotic Operating System (ROS).

\begin{figure*}[t]
    \centering
    \includegraphics[width=0.7\linewidth]{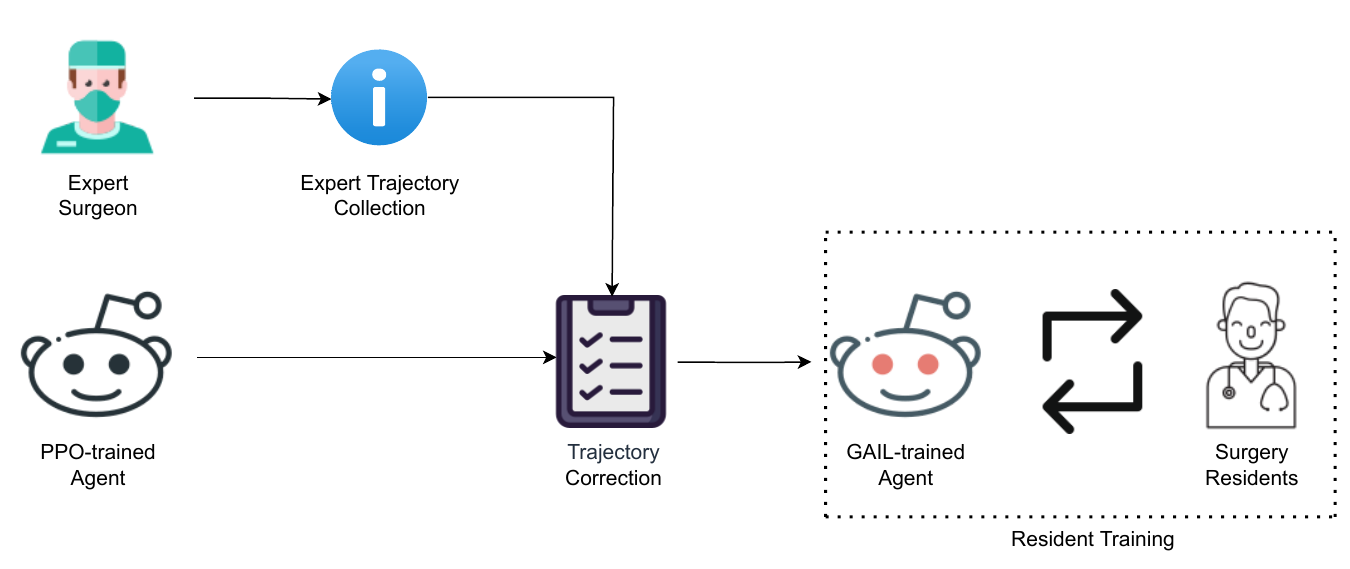}
    \caption{Pipeline implemented by Tan et al.~\cite{tan_laparoscopy} to collect data from an expert surgeon to train a GAIL agent. The agent is then tasked with on-training surgical residents through practical demonstrations. Trajectory correction is performed to account for proper handling motion to improve surgeon precision and reduce fatigue}
    \label{fig:gail_laparoscopy_flow_chart}
    \Description{Laparoscopy GAIL Agent Pipeline shows data collected from expert surgeon to train a GAIL agent, then the agent performs practical demonstrations for surgical residents on-training.}
\end{figure*}

\subsubsection{Actor-Critic with Demonstrations:}
Following a somewhat similar formula of learning from demonstrations, Keller et al.~\cite{keller_dalk} uses the DDPGfD framework~\cite{DDPGfD} to automate the incision phase of deep anterior lamellar keratoplasty (DALK) microsurgery using a surgical robot. Since the separation of the cornea layers is considered exceedingly difficult and the difference between the success and failure of the surgery is determined by tens of micrometers \cite{keller_dalk}, automation of the needle insertion process was able to outperform surgical fellows in reaching a precise target needle insertion depth in mock corneal surgery trials.
The approach used recorded demonstrations from expert surgeons stored in the replay buffer as a foundation for the training of the critic network, then improved on it with reinforcement learning.

\subsubsection{Incentivized Exploration:}
On the other hand, since the use of reinforcement learning is still a significant part of these algorithms, the exploration phase of the training is a present issue. In addition, improving exploration efficiency by giving higher sample priority to learning the expert policy over environment exploration adds a new bottleneck: the amount of demonstration data available. Huang et al.~\cite{huang2023demonstrationguided} proposed a novel approach to increase exploration efficiency. Their methodology lies in adding a coefficient to steer the exploration-exploitation balance; then discounting the reward at a specific time step by the distance $d$ (weighted by the coefficient) between the agent action and the expert action at that given state (Equation~\ref{huang_equation} shows the regularized Q-value function).

\begin{equation}
\label{huang_equation}
    Q^{\pi}(s,a) := \EX_\pi\left[\sum_{t=0}^{\infty}\gamma^t(r_t-\gamma\alpha d(a_{t+1}, a_{t+1}^e))\big|s, a \right ]
\end{equation}
This approach, however, incentivizes the agent to explore in the same direction of the expert demonstrations, which makes it highly sensitive to subpar performance in said demonstrations.

\subsubsection{Relevant expert knowledge provided by the surgeon:}

When it comes to what kind of expert knowledge should be provided to the agent, it depends highly on the intended application of the agent itself. Tan et al.~\cite{tan_laparoscopy} performed a data collection phase in their application pipeline in which trajectory data was gathered from experienced senior surgeons. The collected data were then corrected for proper handling motion to teach the agent how to demonstrate the appropriate relaxed posture of the surgeon’s wrist to improve precision and reduce fatigue during the operation (shown in~\autoref{fig:gail_laparoscopy_flow_chart}).
A different application, such as the one proposed by Chi et al.~\cite{chi_catheter}, used data that were actively collected from expert surgeons performing the selective catheterization task on a robotic platform called CathBot, which was introduced in~\cite{cathbot}. The information gathered by the system during demonstrated operations includes:

\begin{itemize}
\label{itm:chi_items}
        \item \textbf{Motion Signals}: An electromagnetic tracking system with six degrees of freedom is attached to the tip of the catheter. The setup aims to gather information based on the following:
        \begin{itemize}
            \item Catheter tip spatial position/rotation
            \item Linear and angular velocity of the CathBot manipulator
        \end{itemize}
        \item \textbf{State-action Pairs}: The different actions performed by the operating surgeon are recorded and discretized into the following primitives:
        \begin{itemize}
            \item \textit{Push}: The operator pushes the catheter tip into the artery
            \item \textit{Pull}: The operator pulls the catheter tip from the artery
            \item \textit{Clockwise rotation}: The operator rotates the catheter in a clockwise motion
            \item \textit{Anti-clockwise rotation}: The operator rotates the catheter in an anti-clockwise motion
            \item \textit{Standby}: The operator does not perform any action
        \end{itemize}
\end{itemize}

\subsubsection{Surgery Procedure Transparency vs. Obscurity:}

Implementing a simulated digital twin for a surgical environment is a highly complicated task~\cite{erol2020digital,ahmed2021potential} prompting the question of whether to include all parts of the actual surgery in the simulation of the digital twin. Attempting to simulate every physical and biological aspect of what happens during actual surgery can skyrocket computational costs, with frequently diminishing improvements. However, various proposals successfully trained an agent in a simplified simulation. In the approach proposed by Chi et al.~\cite{chi_catheter}, the GAIL agent was trained to successfully perform selective catheterization in an environment that only included soft vascular phantoms, blood flow simulation, and off-the-shelf surgical instruments. In the proposal by Tan et al.\cite{tan_laparoscopy}, the built-in functionalities of the Virtual Robot Experimentation Platform (V-REP) are used to simulate the usage of the laparoscopy tool.

\bgroup
\def\arraystretch{1.2}
\begin{table*}
\centering
\resizebox{\linewidth}{!}{
\begin{tabular}{|p{0.1\linewidth}|p{0.26\linewidth}|p{0.26\linewidth}|p{0.26\linewidth}|}
 \hline
 \textbf{Approach (Type of surgery)} & \textbf{State representation} & \textbf{Spatial representation} & \textbf{Action representation}\\
 \hline  \hline
 Tan et al.~\cite{tan_laparoscopy} (Laparoscopy) & Divide the operation into a set of trajectories in which surgical tools are manipulated along a set path & Mask R-CNN is
used to segment and track laparoscopic tools from visual input. Expert trajectories utilize various joint positions stored on FPGA resources & Linear and angular transformation of the right-hand motion of the tool depending on the task trajectory \\
 \hline
 Chi et al.~\cite{chi_catheter} (Cathertization) & Relative distance between the current spatial position and the position of the target vessel & Positional and rotational information of the catheter tip + Linear and angular velocities of the manipulator tool (as seen in~\ref{itm:chi_items}) & Five action primitives: Push, Pull, Rotate Clockwise, Rotate Anti-clockwise, and Standby (Shown in~\ref{itm:chi_items}) \\
 Li et al.~\cite{li_laparoscopy} (Laparoscopy) & The state space consists of the visual input itself, an RGB-D observation rendered by the simulation system. & Tool position $(x, y, z)$ and orientation $(\alpha, \beta, \gamma)$ relative to current camera pose & Tool transformation vector $(x, y, z, \alpha, \beta, \gamma)$ relative to current camera pose \\
 \hline
 Kim et al.~\cite{Kim2021_il} (Eye Surgery) & A sequence of Euclidean points (3D points) on the surface of the retina inside the eye to which the tool must be navigated. The state is defined as the tool position relative to the next destination & Represented as $x = (p, R, \upsilon, \omega)$ where $p\in \mathbb{R}^3$ defines the tooltip position, $R \in SO(3)$ the orientation matrix, $ \upsilon \in \mathbb{R}^3 $ the linear velocity of the tool frame and $ \omega \in \mathbb{R}^3 $ the angular velocity of the tooltip & The tool is navigated from the current position to the target position in a straight line and makes very gentle contact with the retina. The microscope feed and tooltip position based on robot kinematics are used as the expert action space \\
 \hline
 Keller et al.~\cite{keller_dalk} (DALK Surgery) & 
 The state space consists of seven values: $\Delta{x}$ and $\Delta{y}$ from the goal, $\Delta$pitch and $\Delta$yaw to the face goal, needle \% depth, $\Delta$depth from goal and corneal deformation value $d$ equal to the sum of squared residuals of a second order polynomial fit to the epithelium & Represented as spatial coordinates $x,y,z$ as well as the pitch and yaw of the needle tip & 
Represented as yaw and pitch changes of $\pm 5 \degree$ 
(clamped between $-20\degree$ to $20\degree$ yaw and $-5\degree$ to $25\degree$ pitch) and movement of the needle between 10 to 250 $\mu$m in the direction it was facing at each time step \\
 \hline
 Huang et al.~\cite{huang2023demonstrationguided} (General Surgery) & 
 The low-dimensional state representation used consists of the target object state (3D Cartesian positions and 6D pose) and the robot's spatial and motion state (jaw status and end-effector position)
& 
Utilized as-is in the state representation. Consists of $x,y,z$ coordinates and additional orientation degrees (pitch and yaw) & 
Represented as $(d_x,d_y,d_z,d_{yaw}/d_{pitch},j)$ where $d_x,d_y,d_z$ determine the translation vector in Cartesian space, $d_{yaw}/d_{pitch}$ determines the orientation vector, and $j$ determines whether the jaw is open $(j \ge 0)$ or closed $(j < 0)$ \\
 \hline
 Segato et al.~\cite{segato_keyhole} (Keyhole Neurosurgery) & 
 Represented as a set of observations collected from the environment. These include the cumulative step count $t$, whether an obstacle collision occurred, whether the target was reached (degree of tolerance $tol=2.5mm$), and the distance to target position & 
Agent spatial configuration is represented as a $4 \times 4$ pose matrix:
$\begin{pmatrix}
    R(q_{needle}) & p(q_{needle}) \\
    0^T & 1
\end{pmatrix}$
 where the orientation and position of the agent are denoted as $R(q_{needle}) = \big[\alpha,\beta,\gamma\big]^T$ and $p(q_{needle}) = \big[x,y,z\big]^T $ respectively.  & 
Represented at time step $t$ as $a_t = \big[x,y,z,\alpha,\beta,\gamma\big]$ where the agent can use a combination of the translation along its $z$ axis and the rotation about both $x$ and $y$ axes due to the RCM constraint \\
 
 \hline
\end{tabular}
}
\caption{Different methods for state, spatial, and action representation of the expert knowledge.}
\label{table:relevant_knowledge}
\end{table*}
\egroup

\section{Discussion and Outlook}

Several problems are typically encountered when designing an AI-based surgical robotic assistant. As discussed earlier, while introducing human-in-the-loop learning alleviates some of these problems, it comes with its own challenges. We highlight these challenges in a more pronounced way on the basis of our survey and then propose possible future research directions.

\subsection{Challenges in Current Robotic Surgery Approaches}

\subsubsection{Identifying the relevant knowledge:}

As mentioned above, the relevant expert knowledge that facilitates high-performance systems highly depends on the intended purpose. Defining the amount and type of information to include can prove challenging in some contexts. However, regardless of the method and type of data collected, critical information usually contains a discretization of the operating states~\cite{tan_laparoscopy,chi_catheter,li_laparoscopy}, a spatial representation of the linear/rotary motions of the actuator, and the appropriate action representation to reach the desired state. \autoref{table:relevant_knowledge} showcases different methods used to incorporate the relevant data needed by the hybrid model.

\subsubsection{Medical knowledge representation:}

Converting knowledge in the field of medicine to a machine-understandable format is often quite challenging~\cite{kononenko1997application, musen1992dimensions, koutsojannismedical}. In medicine, the primary goal is the care provided for a specific patient, while dealing with patients as a research resource is a secondary nonnuanced goal. As for surgery, the secondary goal is even less essential and surgeries require critically high safety standards and incredibly low error margins. While machines can be even more consistent and accurate than human surgeons, as discussed earlier, bridging the gap between the cumulative knowledge of experts and the mathematical representations of such data becomes a clear obstacle for the system designer, with neither having sufficient knowledge nor enough help from surgeons.

\subsubsection{Designing the proper transparency and interpretability level:}

Clarity is essential in medicine, especially surgery, as information from various sources must be evaluated to make clinical decisions. Most deep learning approaches incorporate a ``black box'' approach in which the user has a severe limitation on interpreting, understanding, or correcting how a model has made its decision. Human-in-the-loop learning allows medical experts to thoroughly evaluate, assess, and nurse the AI agents in their fields, which eases the correction of the black-box approach. However, interpretability and transparency remain a difficult challenge~\cite{chakraborty2017interpretability, stoyanov2018understanding, holzinger2017towards}.

\subsubsection{Acquiring the expert knowledge:}

The data collection task in most AI-based use cases in the context of medicine is rarely trivial. Furthermore, there is no shortage of proposed methods and tools to collect tabular, statistical, and numerical data~\cite{jansen2005guidelines, nelson1998building, rolim2010cloud, cardiocube, xu2009unsupervised}, human-in-the-loop learning relies on expert knowledge to achieve its potential. Additionally, surgery, in particular, relative to most other fields of medicine, is regarded a highly complicated practice, with surgical residents undergoing 5 to 8 years of residency compared to 3 to 4 years for physicians~\cite{length2022residencies}, making the acquisition of surgical expertise both costly and tedious~\cite{zakim2015data}.

\subsection{User-centered design for Robotic Surgery}

The previously discussed challenges emerge from the fact that the current approaches, while considering human-in-the-loop, focus on a one-model-fits-all approach where the robot agent has to learn from multiple surgical experts and implicitly consolidate their behavior and actions. We suggest a more user-centered approach that considers each surgical expert's knowledge separately in a unique and personalized manner. User-centered adaptive interfaces~\cite{rogers2000adaptive,hassel2005adaptation,janarthanam2014adaptive,manawadu2017multimodal,neverova2015moddrop,gnjatovic2012adaptive,gomaa2021ml} have been investigated with multiple domains (e.g., automotive and robotics) for multimodal interaction using speech, gaze, and gesture. Although surgery is not considered to be an interaction with an interface, the surgeon would still have to interact with the robotic surgeon until fully autonomous agents are reached in the future. Thus, we look at these adaptation and personalization techniques as inspiration.

For example, transfer learning~\cite{da2019survey} approaches could be utilized in fine-tuning the learning of a specific task from a generalized one by training on the standard surgical procedure from multiple different surgical specialties (e.g., Laparoscopy and Catheterization) then fine-tuning on an entirely different surgical procedure with similar initial steps (e.g., Eye Surgery). This would help solve the challenges of both \textit{ Identifying the relevant knowledge} and \textit{Medical knowledge representation} as we use pre-represented data from different surgical procedures and reuse existing data distributed on different entities. However, a community effort in terms of open science and reproducible work must be governed and maintained to achieve this goal.

Similarly, a personalized model can be achieved for each surgical expert by adapting the model based on user-centered data either by fine-tuning or by collecting data solely from this expert surgeon. While this approach will not constitute a high degree of external validity, it will maintain a high internal validity with high user-specific performance. This will contribute to solving the \textit{Designing the proper transparency and interpretability level} challenge as it will be highly customized to each surgeon, promoting trust in the system. Furthermore, it will help to overcome the \textit{Acquiring the expert knowledge} challenge as it will rely on a few surgical experts only that provide a large amount of information instead of the need to have many surgeons, each of whom provides a limited amount of information to the system.

Although the approaches previously suggested manipulate the data collection step to overcome implementation challenges, another approach would be to enhance the learning models by looking at incremental~\cite{gepperth2016incremental}, online~\cite{guo2022online}, and feedback-based learning~\cite{knox2009interactively} domains. More specifically, while most of the previously mentioned learning techniques insert prior knowledge in an offline predetermined manner, online feedback-based learning approaches could be utilized to directly teach the robot agent how to perform the surgery accurately. As an example, this could be achieved by combining imitation learning techniques with online interactive methods such as explicit feedback (e.g., TAMER~\cite{knox2009interactively}) or implicit feedback (e.g., EMPATHIC~\cite{cui2020empathic}) approaches.

\section{Conclusion}

Several decades ago, surgical robots were an implausible idea that could never be achieved; however, the da Vinci Surgical Robotic System~\cite{dimaio2011vinci} has been assisting surgeons to perform millions of operations for the last ten years~\cite{intuitive2022Davinci}. Similarly, with the current advancement in artificial intelligence and machine learning, fully autonomous surgical robots could be a plausible, feasible, and achievable concept. Motivated by this goal, we have surveyed several hybrid machine learning techniques considering the human-in-the-loop paradigm to achieve a robot teaching environment. We highlight some existing approaches for AI-based robotic surgeons that can implement simple steps in a surgical procedure. In addition, we extract and illustrate the current challenges in these systems from a theoretical and practical point of view and highlight possible challenges in the future system. Inspired by transfer learning techniques, we propose a guideline for user-centered design for robotic surgeons that could remedy some of these challenges and hopefully inspire future researchers to achieve their goal of creating these systems. 

\begin{acks}

This work is partially funded by the German Ministry of Education and Research (BMBF) under the TeachTAM project (Grant Number: 01IS17043).

\end{acks}

\bibliographystyle{ACM-Reference-Format}
\bibliography{references}


\end{document}